\documentclass[runningheads]{llncs}
\usepackage[T1]{fontenc}
\usepackage{graphicx}
\usepackage{booktabs}
\usepackage[misc]{ifsym}
\newcommand{\corr}{(\Letter)}
\usepackage{mwe}
\usepackage{amsmath}
\usepackage{amsfonts}
\usepackage{booktabs}
\usepackage{pifont}
\usepackage{makecell}
\newcommand{\cmark}{\ding{51}} 
\newcommand{\xmark}{\ding{55}} 

\begin{document}

\title{PRISM-CTG: A Foundation Model for Cardiotocography Analysis with Multi-View SSL}

\titlerunning{PRISM-CTG: A Foundation Model for CTG Analysis with Multi-View SSL}

\author{Author information scrubbed for double-blind reviewing}
\author{Sheng Wong\inst{1,2} \corr \and
Ravi Shankar \inst{1,2} \and
Beth Albert \inst{1,2} \and
Hao Fei \inst{3} \and
Lin Li \inst{3} \and
Imane Ben M'Barek \inst{4} \and
Manu Vatish \inst{1} \and
Gabriel Davis Jones \inst{1,2}
}

\authorrunning{S.F. Wong et al.}

\institute{Nuffield Department of Women's \& Reproductive Health, University of Oxford, United Kingdom
\and
OXDHL, University of Oxford, Oxford, United Kingdom
\and
OATML, Department of Computer Science, University of Oxford, United Kingdom
\and
Université Paris Cité, Paris, France
}
\maketitle              

\begin{abstract}
Supervised deep learning models for automated CTG analysis are typically constrained by narrowly curated labelled datasets and limited patient cohorts, leaving substantial volumes of physiologically informative clinical recordings untapped. To address this limitation, we propose Physiology-aware Representation Learning via Integrated Self-supervision and Metadata for CTG (PRISM-CTG), a clinically grounded self-supervised foundation model (FM) for CTG that leverages large-scale unlabelled recordings to learn transferable domain-level representations. PRISM-CTG is pretrained using a multi-view self-supervised framework that jointly optimises 3 complementary pretext objectives: random-projected guided masked signal reconstruction, clinical variable prediction, and feature classification. Each objective is associated with a dedicated task-specific token, enabling specialised representation learning, while controlled cross-attention facilitates information exchange across clinical context. By reframing patient metadata and domain knowledge, which are often underutilised in conventional training as prediction targets, Prism-CTG transforms readily available clinical information into additional supervisory targets that guide clinically meaningful representation learning. Extensive experiments across 7 downstream CTG tasks in both antepartum and intrapartum domains demonstrated that PRISM-CTG consistently outperforms in-domain and SSL baselines. Notably, PRISM-CTG demonstrated strong generalisation under external validation on 2 datasets, while achieving comparable performance to studies trained on substantially larger, privately labelled datasets. To our knowledge, this is the first study to introduce large-scale FM for CTG that learns domain-level representations.

\keywords{CTG Analysis \and Self-supervised Learning \and Foundation Model}

\end{abstract}

\section{Introduction}
Cardiotocography (CTG) provides indirect measures of fetal brain activity. It comprises simultaneous time-series recordings of fetal heart rate (FHR) and uterine activity (UA), and remains the primary clinical tool for continuous, non-invasive fetal health assessment during pregnancy and labour~\cite{jones2022computerized}. Despite its widespread adoption, interpreting CTGs is both time-consuming and inherently subjective, requiring constant attention over prolonged monitoring periods~\cite{hernandez2023reliability}. This creates diagnostic uncertainty and could potentially lead to delayed or suboptimal decision-making. In response, deep learning (DL) models have been proposed for automated CTG analysis to assist clinicians in CTG interpretation. However, the performance of these models remains constrained by limited patient cohorts and narrowly curated datasets designed for predefined tasks~\cite{khan2025patchctg,multiscaleLSTM,TGLCN,DWT-CNNBiLSTM}. 

These supervised DL models rely on large quantities of high-quality, clearly defined labels to learn task-specific predictions, with labels typically derived from neonatal outcomes or clinical criteria. In real-world settings, however, many CTG recordings do not align cleanly into these label categories due to several factors, including early recordings that do not reflect final neonatal status, borderline or ambiguous cases, or recordings with significant dropout~\cite{davis2026identifying,khan2025patchctg}. Consequently, a large volume of physiologically informative CTGs is excluded from the training pipeline despite containing meaningful physiological patterns. This results in models that are trained on highly selective subsets of CTG recordings, leading to task-specific representations that lack robustness even within their respective predefined tasks (see supplementary material S1). 

\begin{figure}[!t]
\includegraphics[width=\textwidth]{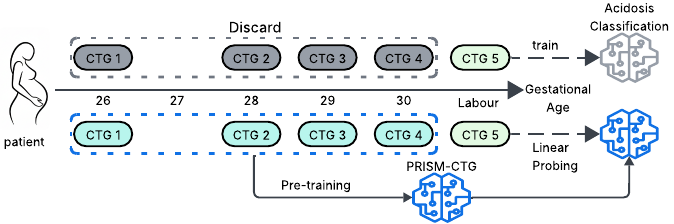}
\caption{Overview of a major limitation in conventional supervised CTG training. Supervised models typically select temporally aligned recordings for a single predefined task, discarding CTGs acquired earlier in the gestational period (top). In contrast, PRISM-CTG leverages CTGs from the full gestational period during pretraining to learn general representations that are transferable across multiple downstream tasks.} 
\label{problem statement}
\end{figure}

Pretraining via self-supervised learning (SSL) \cite{mae,jepa} offers an alternative for learning meaningful representation from unlabelled data and has demonstrated strong performance in the physiological signals domain, enabling the development of foundation models (FMs) that are capable of multiple downstream tasks in their respective domain \cite{eegformer,VQMTM,papagei,alfee}. However, SSL-based FMs have not been investigated in the domain of CTGs. Additionally, existing SSL approaches adopted by these FMs predominantly rely on single signal pretext objectives, which, while effective at capturing low-level signal characteristics, largely overlook available clinical context. In practice, clinical context, including patient metadata and domain-engineered features, is often ignored or discarded, despite encoding clinically meaningful physiological and contextual information~\cite{davis2026identifying,ben2023computerized,jones2022computerized}. Although individually insufficient for diagnosis, this information reflects expert knowledge routinely used in clinical interpretation and can provide valuable inductive bias during SSL~\cite{davis2026identifying,ben2023computerized,jones2022computerized}.

To this end, this work presents \textbf{PRISM-CTG}, a FM for CTG analysis that learns clinically grounded, transferable CTG representations. Prism-CTG is pretrained on over 250,000 hours of CTGs collected across 3 decades using a multiview SSL framework that leverages 3 complementary supervisory signals: random-projected guided masked signal reconstruction (Best-RQ-MAE) to learn general CTG patterns, patient metadata prediction, and prediction of domain-engineered features. By reframing otherwise discarded clinical information as free supervisory targets, PRISM-CTG integrates signal-level and patient-level context through dedicated task-specific tokens with controlled cross-attention. This design mirrors clinical CTG interpretation, where signal inputs are evaluated alongside patient context, enabling the model to learn context-aware representations that are robust and transferable across downstream tasks. In summary, the main contributions of this study are \textbf{the first CTG FM} pretrained to learn transferable domain-level representations by integrating morphology-guided reconstruction with clinically informed supervision. \textbf{Multi-view objective with task-specific representation specialisation} through isolated task tokens and controlled cross-attention that integrates signal-level, patient-level, and feature-level supervision during pretraining. PRISM-CTG is \textbf{transferable} across antepartum and intrapartum domains, achieving an average improvement of 4.93\% to 19.31\% over supervised models. PRISM-CTG demonstrated strong \textbf{cross-institution generalisation}, achieving average improvements of 2.76\% and 8.62\% on two external datasets, despite minimal intrapartum data during pretraining. 
    

\section{Related Work}
\subsection{CTG Analysis}
Recent advances in automated CTG analysis have largely been driven by supervised learning approaches~\cite{neurofetalnet,khan2025patchctg,CTG-net,multiscaleLSTM,TGLCN,deepctg}. These models are typically trained explicitly on single clinical objectives such as fetal acidemia detection or abnormality classification, relying on very small volumes of CTG recordings. In an attempt to address label scarcity, some studies have explored supervised pretraining strategies to improve on the task performance~\cite{CTG-net,khan2025patchctg,deepctg} (See supplementary material S2 for more details). Although these methods improve optimisation stability and task-specific performance, they were not aimed to learn general CTG representations, only task-specific representations. More recently, a small number of works in the field of CTG analysis have incorporated SSL techniques~\cite{ctggan,mctg,9747178}. However, these approaches are still typically embedded within a task-specific pipeline, where the focus was not on large-scale pretraining. In summary, previous work has largely focused on task-specific optimisation with pretraining, instead of learning transferable, general representations. PRISM-CTG addresses this gap through large-scale SSL, learning a robust general representation that is transferable to multiple downstream tasks. 

\subsection{Foundation Models for Physiological Time-series}
Self-supervised pretraining for physiological time series is well established. They are often adapted from other vision or time-series domains to build domain-specific FMs \cite{VQMTM,esi,sleepfm,MELPS,ZeroBurden,alfee,labram,eegformer,papagei,wildecg}. Trained on large in-domain physiological data, these FMs have demonstrated transferability across multiple downstream tasks. However, these domain-specific approaches, along with general-purpose time-series SSL methods, do not transfer well to CTG analysis. Many rely on assumptions that do not hold for CTG signals. For example, some studies rely on inter-channel or cross-signal representation alignment, an approach well suited to multimodal physiological settings where signals are recorded synchronously and reflect shared underlying neural or cardiac activities~\cite {anyppg,sleepfm}. Other FMs utilised signal transformations to derive domain-informative representations during pretraining, including spectral and frequency-domain representations, to guide representation learning~\cite{VQMTM,eegformer,alfee}. In contrast, CTG comprises two physiologically distinct signals. FHR is a non-periodic signal reflecting autonomic regulation of the fetus, while UA originates from the maternal uterine contractions that exhibits periodicity and temporal frequency that varies with the stage of pregnancy. As such, we argue that these methods are not well-suited to learning robust representations from CTG signals. 

\subsection{Pre-training with auxiliary information}
Traditionally, SSL relies solely on the target modality during pretraining. Some FMs in this domain extend beyond signal-only SSL by incorporating auxiliary inputs or targets such as textual reports or engineered features to guide representation learning. Several domain-specific FMs employ cross-modal alignment strategies with paired signal and text data to learn semantically transferable embeddings~\cite{esi,MELPS,tolerantecg}. Other approaches adopt predictive pretraining paradigms that use auxiliary signals or engineered features supervision signals to guide representation learning~\cite{hubert,clef,ZeroBurden}, relying on strong domain knowledge as pretraining objectives, resulting in representations that reflect in-domain physiological or clinical-features characteristics. Motivated by these methods, we incorporated patients' metadata and engineered features directly into the pretraining objectives to guide representation learning within a predictive paradigm, without relying on additional signal modalities.

\section{Methods}

\begin{figure}[t]
\includegraphics[width=\textwidth]{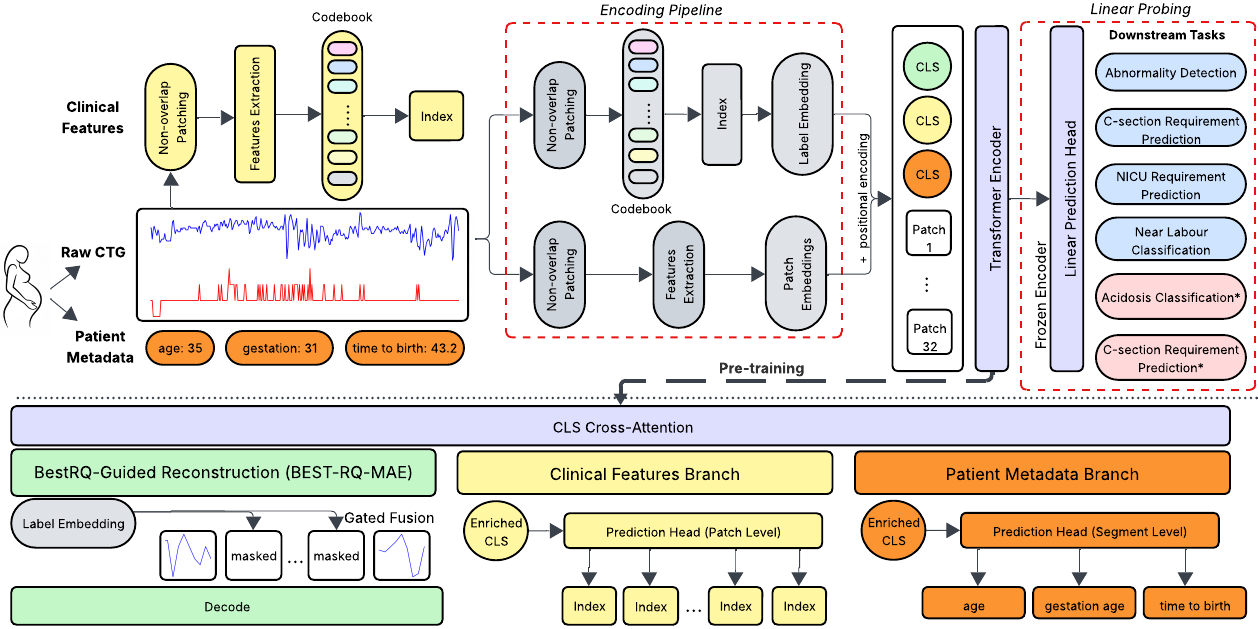}
\caption{PRISM-CTG architecture. The model is pretrained with 3 pretext objectives and evaluated via linear probing by freezing the encoder and training a task-specific linear head on 6 downstream tasks (four antepartum, two intrapartum*).} \label{fig1}
\end{figure}

In this section, we describe the overall architecture of PRISM-CTG, specifically highlighting the modules that are unique from standard SSL. We use the terms task token and \texttt{CLS} token interchangeably. A CTG recording is represented as a multivariate time series $\mathbf{X}\in\mathbb{R}^{L\times C}$, where $C{=}2$ (FHR and UC) and $L$ is timepoints. The model jointly optimises 3 complementary pretext objectives: (i) Best-RQ-MAE, (ii) patient metadata prediction, and (iii) domain feature label classification.  We assign 3 task-specific tokens, where they independently aggregate information from the input during encoding via an isolation attention mask. This enforces objective-specific specialisation from shared patch tokens. After encoding, we apply dedicated \texttt{CLS} cross-attention blocks, allowing controlled information exchange only after each task token has formed its own summary. Notably, in our proposed Best-RQ-MAE module, a parallel frozen random-projection tokeniser assigns discrete pseudo-labels to raw patches, and the resulting label embeddings are injected into both encoder and decoder to provide morphology-aware, signal-level guidance during reconstruction. In the decoder, label embeddings are fused with patch representations through a learned gate, and patch tokens are further conditioned on task-specific CLS summaries via patch-to-CLS cross-attention. The 3 objectives are jointly optimised with learned uncertainty-based weighting to balance heterogeneous scaling losses. (see supplementary material S3 for the full configurations and details).

\subsection{Encoding Pipeline}
Given a CTG recording $\mathbf{X}\in\mathbb{R}^{L\times C}$ ($C{=}2$: FHR and UC), Prism-CTG builds patch tokens through two parallel streams: a learned CNN patch embedding and a frozen random-projection label embedding.

\subsubsection{CNN Backbone and Patch Embedding}
The CNN backbone, which consists of multiple residual convolutional blocks, extracts local features from the raw CTG signal without temporal downsampling. The resulting feature maps are partitioned into non-overlapping patches using a strided 1D convolution. Each patch is then linearly projected to an embedding dimension $D$, producing $\mathbf{h}_i$.

\subsubsection{Signal Tokenisation and Label Embedding}
In parallel with patch embedding, we adopt a frozen random-projection quantizer, inspired by BEST-RQ to assign each patch a discrete pseudo-label \cite{bestrq}. Specifically, non-overlapping patches are first projected into a fixed latent space using a random projection and matched to the nearest neighbour in a frozen codebook, producing discrete indices that act as pseudo-labels. They are then mapped into $D$-dimensional embedding vectors via a learnable embedding table, forming label embeddings, $\mathbf{e}_{\ell_i}$. Importantly, the novelty does not lie in the quantisation itself, but in how the pseudo-labels are integrated into the pretraining framework. By grouping patches with similar characteristics, these embedding labels projected from pseudo-labels provide signal-level structural guidance during representation learning.

\subsubsection{Patch Fusion}
For each patch position $i$, the encoder input token is formed by element-wise addition of 3$D$-dimensional components: $\mathbf{z}_i=\mathbf{h}_i+\mathbf{e}_{\ell_i}+\mathbf{p}^{\text{pos}}_i$
where $\mathbf{h}_i$ is the CNN-derived patch embedding, $\mathbf{e}_{\ell_i}$ is the label embedding and $\mathbf{p}^{\text{pos}}_i$ represents positional encoding. Together, the components serve as the input representation for Transformer encoding.

\subsection{Clinical-Aware Transfromer Encoding}
Conventionally, transformers employ a single \texttt{CLS} token to capture global information for downstream tasks. To enable clinical-aware representation learning, we instead prepend 3 learnable task tokens to the patch-token sequence:
\(\texttt{CLS}_r\) (BestRQ-MAE), \(\texttt{CLS}_v\) (patient metadata prediction), and \(\texttt{CLS}_f\) (feature classification). We explain the architecture design choice below. 

\subsubsection{CLS Isolation Attention Mask}
Rather than allowing all tokens to attend freely, we enforce an isolation constraint through attention masking. Each \texttt{CLS} token attends only to itself and to all patch tokens. This forces each \texttt{CLS} token to independently aggregate information from the patches, developing a specialised summary controlled by its respective task objective. After encoding, the output is partitioned into the three enriched \texttt{CLS} tokens and the remaining patch embeddings, defined as $\hat{\mathbf{H}} \in \mathbb{R}^{N \times D}$.

\subsubsection{Task-Wise Cross-Attention}
After Transformer encoding with the isolation mask, each \texttt{CLS} token contains task-specific summaries. We then introduce cross-attention to enable controlled information exchange between \texttt{CLS} tokens, where for each task token, we compute $\mathbf{cls}_{task}' = \mathrm{CrossAttn}\!\left( \mathbf{cls}_{task},
[\mathbf{cls}_r;\, \mathbf{cls}_v;\, \mathbf{cls}_f]
\right)$, and $task \in \{r, v, f\}$. From a clinical perspective, CTG interpretation is inherently multi-factorial: signal characteristics, patient context, and clinical features inform one another. For example, the interpretation of a CTG may depend on the baseline and gestational age. The cross-attention component reflects this controlled interaction by enabling each objective to adaptively integrate relevant clinical information while preserving task-specific structure. 

\subsection{Multi-Task Pretext Objectives}
The model is jointly trained on 3 complementary pretext objectives that operate at different levels of granularity, each utilising a dedicated enriched \texttt{CLS} token. Best-RQ-MAE encourages the learning of long-range temporal dependencies and signal characteristics, patient metadata prediction provides global recording-level supervision grounded in patient context, and feature classification enforces local patch-level representations that align with domain knowledge.

\subsubsection{Best-RQ-MAE}
For the reconstruction objective, we adopt an MAE-style pretraining strategy in which masked patches are reconstructed from visible patches using a lightweight decoder. The key design change lies in introducing morphology-aware conditioning within the decoder through label embeddings derived from the frozen signal tokeniser. Since the tokeniser groups patches with similar signal morphology or patterns, the resulting label embeddings capture the general structure of a CTG signal, providing structured guidance during reconstruction. We compute a gating vector $\mathbf{g}_i = \sigma\!\left(W_g[\mathbf{x}_i;\mathbf{e}_{\ell_i}]\right)$ and fuse the signal guidance as $\tilde{\mathbf{x}}_i = \mathbf{x}_i + \mathbf{g}_i \odot \mathbf{e}_{\ell_i}$, where $\mathbf{x}i$ denotes the restored representation at position $i$, and $\mathbf{e}{\ell_i}$ is the corresponding label embedding computed from the \emph{full unmasked} signal. In addition, we inject clinical aware information, \texttt{CLS}, through cross-attention. Specifically, each restored decoder representation queries the enriched reconstruction token $\mathbf{cls}_r'$, enabling incorporation of clinically informed global context before final reconstruction. The reconstructed raw patches are then supervised using masked-patch mean squared error (MSE).

The rationale behind this design is to maintain a lightweight decoder while compensating with structured guidance, avoiding the problem identified by~\cite{Pixio} where a shallow decoder forces the encoder to sacrifice capacity for low-level reconstruction at the expense of semantic understanding. While deeper decoders can improve reconstruction ability, they introduce additional computational overhead \cite{Pixio}. Instead, we provide the decoder with 'morphological hints' derived from the frozen Best-RQ tokeniser, together with additional clinical information from the CLS, allowing the lightweight decoder to reconstruct effectively without requiring additional depth.

\subsubsection{Patient Metadata Prediction}
Clinical interpretation of CTG is context-dependent, where similar morphology patterns may carry different clinical meanings under different patient conditions. To encode the clinical information as global context into the learned representation, we use the enriched token \(\mathbf{cls}_v'\) with a lightweight head that predicts \(d_v\) continuous patient metadata. This objective is trained as regression with MSE. This task encourages the model to learn recording-level, clinically contextual representations rather than relying solely on morphological patterns. In this study, the 3 patient metadata used are gestational age, time to birth, and maternal age. (See supplementary materials S3.2.)

\subsubsection{Engineered Features Classification}
Clinical interpretation of CTG signals relies on established domain-relevant features such as baseline FHR and short-term variability, which reflect underlying autonomic regulation of the fetus and are central to deciding neonatal status. As such, these features can serve as supervisory targets to guide representation learning toward clinically relevant signal characteristics. To align learned patch representations with domain knowledge used in clinical CTG assessment, we introduce a patch-level feature prediction objective. For each visible raw patch, we extract 17 handcrafted clinically relevant features. They are then converted into discrete pseudo-labels using a second frozen random-projection quantizer (BEST-RQ), independent of the signal tokeniser. This approach reduces a high-dimensional continuous target to a single classification objective, trained with cross-entropy loss. Similarly, we utilise the enriched $\mathbf{cls}_f'$ with a lightweight classification head to predict the feature pseudo-labels. (See supplementary materials S3.3.)

\subsection{Linear Probing}
For downstream tasks, we use linear probing to assess the performance and transferability of the pretrained representations. The encoder is frozen, and a linear classification head, mapped to the number of classes, is trained. In this study, we concatenate \texttt{CLS} tokens for downstream classification.

\section{Experiments \& Results}
\subsection{Data}
\subsubsection{Preprocessing}
CTG recordings were segmented into 20-minute windows for model training. The signals were downsampled from 4 Hz to 1 Hz, consistent with existing studies~\cite{khan2025patchctg,deepctg}. Training segments with more than 50\% missing data were discarded during pretraining, while downstream evaluation datasets were retained without additional filtering to better reflect real-world clinical settings. (See supplementary material S4.1)

\subsubsection{Pre-training Data}
We curated our pre-training data from two complementary sources covering both antepartum and intrapartum periods. The primary pre-training data is drawn from the Oxford Maternity (OXMAT) database, comprising antepartum CTG recordings along with the corresponding patient's information, collected between 1990 and 2024 from 5 healthcare institutions in the Oxfordshire region \cite{oxmat}. We additionally incorporated the publicly available unannotated intrapartum SPAM CTG dataset \cite{spam}, consisting of 300 intrapartum recordings. The final pretraining dataset comprises 51,336 pregnancies and approximately 753,352 segments. This amounts to over 250,000 hours of training data. (See supplementary material S4.2)

\subsubsection{Downstream Datasets}
For antepartum tasks, we curated CTG recordings recorded between 2024 - 2025 from the same healthcare institutions (OXMAT-2025), which do not overlap temporally with the OXMAT database. To evaluate performance beyond both the antepartum setting and our training institutions, we utilise the publicly available CTU-UHB (Czech Republic) \cite{ctuuhb} and the soon-to-be-released APHP-CTG (France) intrapartum dataset, comprising 552 and 450 patients with intrapartum CTG recordings, respectively. Both datasets originate from external institutions, providing a rigorous evaluation of cross-site transferability and generalisability. Notably, CTU-UHB is widely used for benchmarking intrapartum fetal acidosis classification, enabling us to compare PRISM-CTG with existing studies trained with private datasets. (See supplementary material S4.2)

\subsection{Models Comparisons}
\subsubsection{Baselines Models}
We compare PRISM-CTG with domain-specific CTG architectures: NeuroFetalNet~\cite{neurofetalnet}, PatchCTG~\cite{khan2025patchctg}, CTGNet~\cite{CTG-net}, Multiscale-LSTM~\cite{multiscaleLSTM}, DWT-CNNBiLSTM~\cite{DWT-CNNBiLSTM}, TGLCN~\cite{TGLCN}, and general-purpose models: ResNet \cite{resnet}, TimesNet~\cite{timesnet}, and TimeMixer~\cite{timemixer}. All supervised models are trained from scratch on each downstream task and are therefore task-specific. Additionally, we select SSL methods representative of the major pre-training frameworks adopted in physiological FMs: BEST-RQ~\cite{bestrq}, VQ-MTM~\cite{VQMTM}, CLIP-style~\cite{clip}, DINO~\cite{dino}, MAE~\cite{mae}, Pixio~\cite{Pixio}, JEPA~\cite{jepa}, and NEPA~\cite{nepa}. All SSL baselines follow the same pre-training protocol as PRISM-CTG. (see supplementary material S5 for additional details)

\subsubsection{Downstream Tasks \& Evaluation} 
We evaluate on 7 downstream tasks: (T1) abnormal classification, (T2) caesarean delivery prediction, (T3) intensive care admission prediction, (T4) delivery proximity classification, fetal acidosis classification on (T5) CTU-UHB and (T6) APHP, and (T7) caesarean delivery classification. T1 to T4 are antepartum tasks, while the remaining are intrapartum tasks. The downstream dataset was split into an 80/20 train–test ratio with no patient overlap between the training and test sets, and results were averaged over 5 runs. We follow other studies and use Area Under the Curve (AUC) as the primary evaluation metric, with standard deviation (SD) reported in supplementary material S6 Table 4 and 5.

\subsubsection{Ablation Analysis}
To evaluate the contribution of each architectural module and design choice in PRISM-CTG, we conducted multiple ablations. Specifically, we investigate the effect of: (a) Convolution layers (b) Signal Tokenisation and Label Embedding (c) patient metadata prediction and features classification (Multiview), (d) BEST-RQ-MAE, (e) Frequency representation as input, and (f) Uncertainty Weighting.  

\subsection{Results \& Discussions}
In this study, we present PRISM-CTG, a CTG FM pretrained with a multiview SSL framework. PRISM-CTG achieves the best average AUC of 0.808, consistently outperforming all baselines on 6 of 7 tasks, as shown in Table \ref{tab:full_results}, indicating transferability across both antepartum and intrapartum settings. No other SSL methods achieved performance comparable to PRISM-CTG, with the strongest SSL baseline, MAE at 0.773, followed by the supervised NeuroFetalNet model at 0.771. Notably, PRISM-CTG achieves the largest performance gains on T7 and T4 against the best supervised models (5.32\% and 5.10\% improvement), both caesarean delivery classification tasks, suggesting that our pretraining strategy is particularly effective for modelling signal characteristics relevant to delivery method prediction and for identifying high-risk cases. Additionally, CLIP-1, which incorporates clinical context through alignment-based training, performed worse than single-signal pretext objectives, suggesting that alignment-based pretraining with clinical information may not be optimal for CTG representation learning. 
 
\begin{table}[t]
    \centering
    \caption{Comparison of AUC performance across antepartum and intrapartum CTG tasks. The best AUC for each task is shown in \textbf{bold}, and the second-highest is \underline{underlined}.}
    \begin{tabular}{l c c c c c c c c c}
        \toprule
        & & \multicolumn{4}{c}{\textbf{Antepartum}} & \multicolumn{3}{c}{\textbf{Intrapartum}} & \\
        \cmidrule(lr){3-6} \cmidrule(lr){7-9}
        Model & \makecell{Pre-\\training} & T1 & T2 & T3 & T4 & T5 & T6 & T7 & Average \\
        \midrule
        \multicolumn{10}{l}{$\bullet$ \textbf{Supervised Learning}} \\
        \midrule
        ResNet & \xmark  & 0.664 & 0.684 & 0.670 & 0.772 &  0.819 & 0.802 & 0.808 & 0.746 \\
        TimesNet & \xmark  & 0.536 & 0.607 & 0.622 & 0.708 & 0.703 & 0.761 & 0.807 & 0.678 \\
        TimeMixer & \xmark  & 0.640 & 0.668 & 0.663 & 0.768 &  0.829 & \underline{0.810} & 0.865 &  0.749 \\
        CTGNet & \xmark  & 0.629 & 0.631 & 0.628 & 0.711  & 0.781 & 0.768 & 0.849 & 0.714\\
        PatchCTG & \xmark  & 0.659 & 0.610 & 0.649 & 0.628 & 0.767 & 0.774 & 0.782 & 0.696\\
        NeuroFetalNet & \xmark  & 0.670 & 0.720 & 0.713 & 0.784 & 0.857 & 0.796 & 0.856 & 0.771\\  
        MS-LSTM & \xmark  & 0.646 & 0.663 & 0.575 & 0.647  & 0.742 & 0.659 & 0.813 & 0.678\\
        TGLCN & \xmark  & 0.720 & 0.671 & 0.651 & 0.621 &  0.819 & \textbf{0.813} & 0.834 & 0.733\\
        CNN-BiLSTM & \xmark  & 0.700 & 0.660 & 0.620 & 0.664 & 0.852 & 0.769 & 0.814 & 0.726\\
        \midrule
        \multicolumn{10}{l}{$\bullet$ \textbf{Self-Supervised Learning}} \\
        \midrule
        BestRQ & \cmark & 0.533 & 0.646 & 0.635 & 0.675 & 0.738 & 0.657 & 0.780 & 0.666 \\
        VQMTM  & \cmark & 0.656 & 0.592 & 0.604 & 0.670 & 0.720 & 0.740 & 0.761 & 0.678 \\
        CLIP-1 & \cmark & 0.630 & 0.653 & 0.665 & 0.785 & 0.844 & 0.719 & 0.814 & 0.730 \\
        CLIP-2 & \cmark & 0.625 & 0.703 & 0.690 & 0.731 & 0.802 & 0.786 & 0.873 & 0.744 \\
        Dino & \cmark & 0.659 & 0.726 & 0.728 & 0.795 & 0.733 & 0.679 & 0.827 & 0.735 \\
        MAE    & \cmark & 0.648 & \underline{0.729} & \underline{0.731} & 0.797 & 0.850 & 0.768 & 0.890 & \underline{0.773} \\
        Pixio  & \cmark & 0.665 & 0.716 & 0.694 & 0.772 & 0.820 & 0.769 & \underline{0.904} & 0.763 \\
        NEPA   & \cmark & 0.626 & 0.621 & 0.620 & 0.704 & 0.725 & 0.734 & 0.731 & 0.680 \\
        JEPA   & \cmark & \underline{0.731} & 0.726 & 0.715 & \underline{0.802} & \underline{0.873} & 0.685 & 0.820 & 0.765 \\
        PRISM-CTG & \cmark & \textbf{0.743} & \textbf{0.731 }& \textbf{0.753} & \textbf{0.824} & \textbf{0.883} & 0.806 & \textbf{0.911} & \textbf{0.808} \\
        \bottomrule
    \end{tabular}
    \label{tab:full_results}
\end{table}

\subsubsection{Effect of Training Size}
\begin{figure}[t]
    \centering
    \includegraphics[width=1\linewidth]{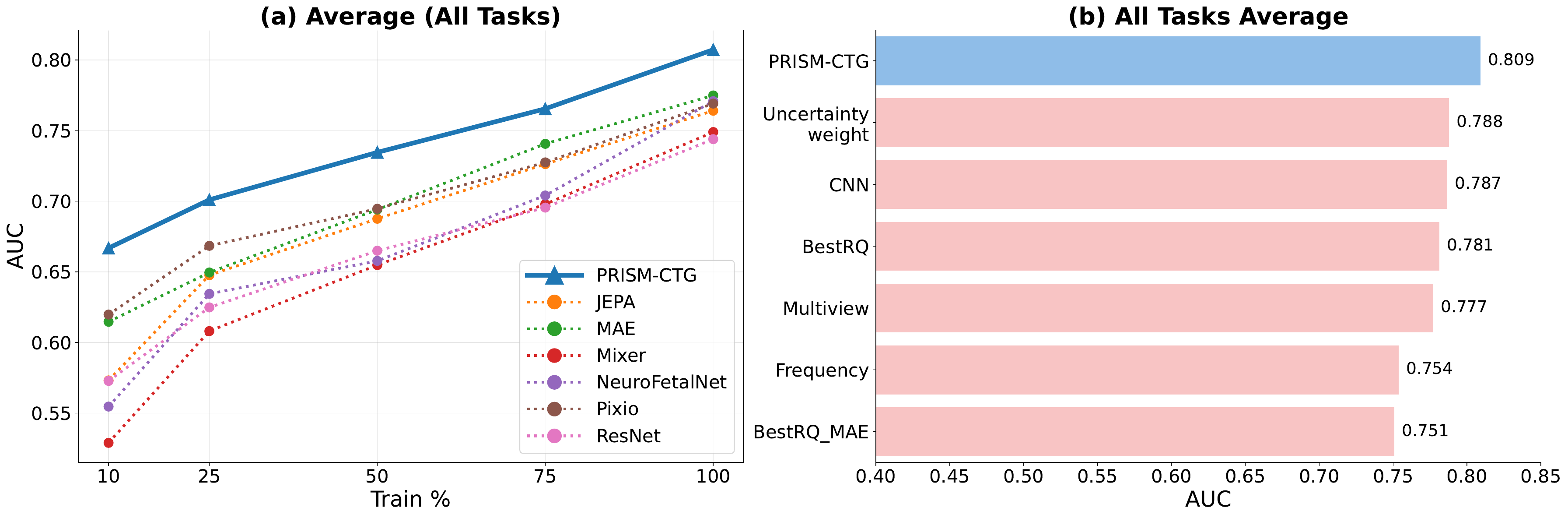}
    \caption{(Left) Average performance across all tasks under different data regimes with the top performing SSL and supervised models. (Right) Ablation study of PRISM-CTG showing average AUC for the full model and its variants.}
    \label{fig:average plot}
\end{figure}

To simulate clinical settings where labelled CTGs are limited or computational resources are constrained, we evaluate PRISM-CTG under varying training data regimes. As shown in Figure~\ref{fig:average plot} (Left), PRISM-CTG maintains strong performance compared with the top baseline models across all data regimes. PRISM-CTG trained with 75\% of labelled data achieves an AUC of 0.765, which is comparable to or exceeds the performance of the top 3 supervised models trained on the full dataset: NeuroFetalNet (0.771), TimeMixer (0.749), and ResNet (0.746). Under the extremely low-data setting (10\% training data), the advantage of pre-training becomes clear. PRISM-CTG outperforms the best supervised baseline, ResNet, with an average improvement of 16.41\%, and the strongest supervised model trained on the full dataset, NeuroFetalNet, at 20.18\%, with even larger gains observed for specific tasks, such as T7 (+30.14\%) under extremely low-data settings. By pretraining on a large volume of CTG recordings to learn a general representation, PRISM-CTG requires fewer data to achieve competitive performance, while the best supervised models remain heavily reliant on larger training data, failing in settings when labelled data are limited.

\subsubsection{Ablation Analaysis}
As seen in Figure 3 (Right), removing the proposed BEST-RQ-MAE module, one of the pretext objectives leads to the largest degradation, reducing average performance by 7.169\%, indicating that BEST-RQ-MAE provides guidance essential for learning a more robust CTG representation. Removing the clinical (multiview) pretext objectives leads to a performance reduction of 3.95\%, suggesting that clinical context helps the model encode global information that guides better representation learning, thereby improving downstream task performance. Additionally, removing label embeddings (BestRQ) resulted in a 3.46\% drop in performance, indicating that signal-level guidance improves representation learning. Finally, we replace the time-domain CTG input with a frequency-domain representation to investigate whether frequency transformations, commonly used in physiological FMs benefit representation learning. However, this results in a 6.80\% decrease in performance, indicating that frequency-domain representations did not provide additional benefit for CTG representation learning.

\begin{figure}[t]
    \centering
    \includegraphics[width=1\linewidth]{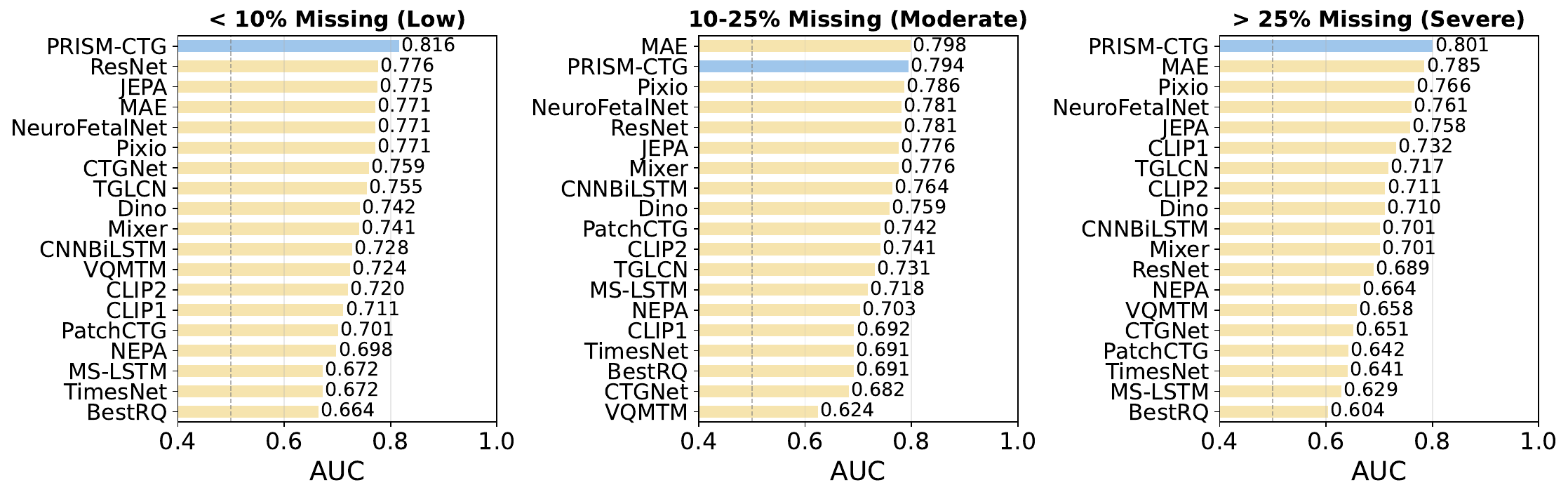}
    \caption{Average performance grouped by proportion of signal dropout}
    \label{fig:missing data}
\end{figure}

\begin{figure}[t]
    \centering
    \includegraphics[width=1\linewidth]{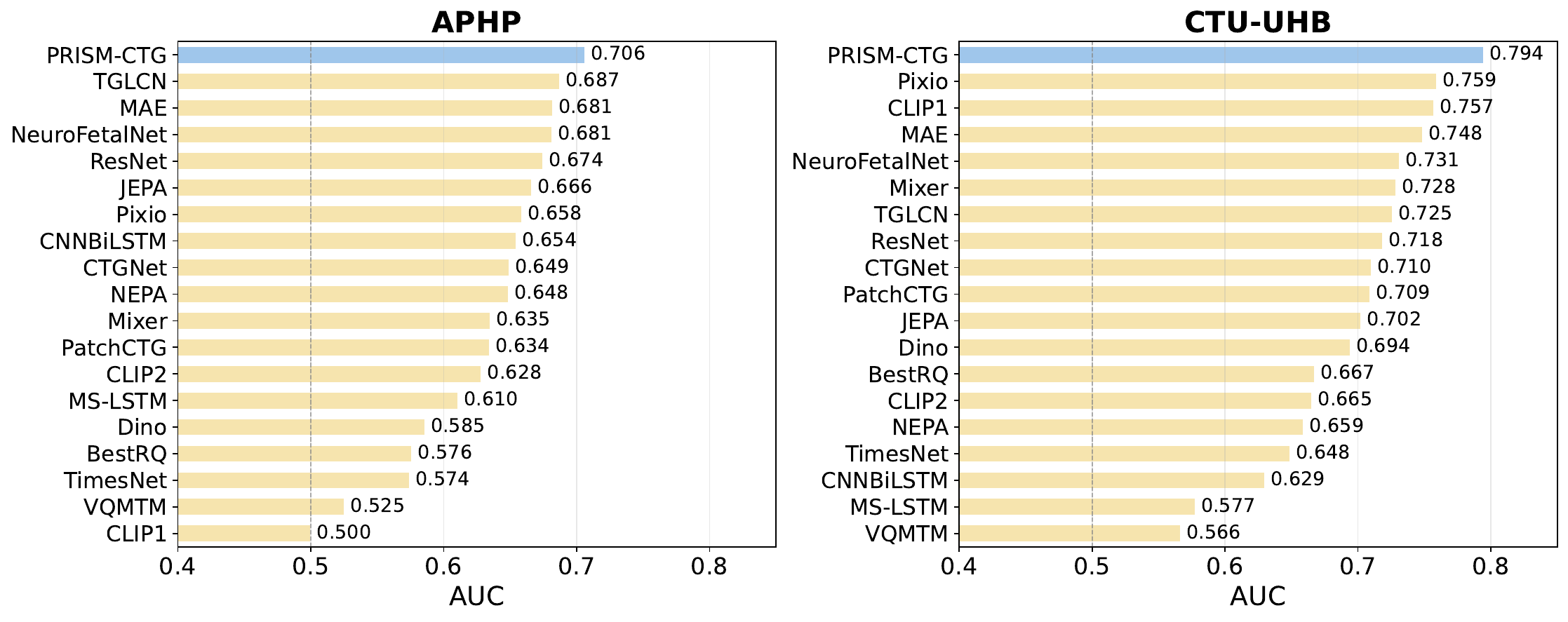}
    \caption{Average performance on external validation on APHP and CTU-UHB.}
    \label{fig:external validation}
\end{figure}

\subsubsection{Impact of Missing Data}
CTG segments are inherently noisy, with signal dropout being one of the most common artefacts, often caused by maternal or fetal movement. To investigate the impact of missingness, we separately evaluate CTG segments grouped according to their proportion of missing data. As seen in Figure \ref{fig:missing data}, We observe that models pretrained via SSL, including PRISM-CTG, demonstrated robustness to signal dropout. In particular, PRISM-CTG and MAE maintain relatively stable performance across increasing levels of signal dropout, while supervised models such as ResNet and TimeMixer experience performance degradation when signal dropout exceeds 25\%. These findings indicate that large-scale SSL pretraining on diverse, real-world CTG recordings improves robustness to signal dropout, resulting in more stable downstream performance under clinically realistic conditions. 

\subsubsection{External Validation}
To demonstrate generalisability, we conducted external validation on the acidosis classification task. Specifically, models were trained on CTU-UHB (T5) and evaluated on APHP-CTG (T6), and vice versa. This cross-dataset evaluation reflects realistic deployment, in which models trained at one institution must generalise to data collected at a different location. As shown in Figure \ref{fig:external validation}, PRISM-CTG remains the best performing model in both tasks and outperforms the best supervised baselines by 2.76\% on APHP-CTG and 8.62\% on CTU-UHB. Notably, intrapartum recordings constitute less than 1\% of the total pretraining data, yet PRISM-CTG maintains superior cross-database performance. We further compared PRISM-CTG with previously published studies trained on large private intrapartum cohorts~\cite{deepctg,mccoy,mcnn}. While our model achieves a lower AUC than some of the highest reported results (AUC = 0.82), those studies were trained on substantially larger datasets, often exceeding 35,000 patients. In contrast, we evaluated our pretrained encoder using linear probing on a cohort of 450 patients (see supplementary material Table 7). Despite the smaller training cohort, performance remains within the range of models trained on private intrapartum cohorts, indicating strong external generalisation while requiring substantially less data. These results indicate that the pretrained encoder captures generalisable CTG representations that transfer effectively across institutions, even when the target domain (intrapartum) is minimally represented during pretraining.

\subsubsection{Limitations}
Although evaluated on the largest available CTG datasets across multiple tasks, the overall data scale remains modest compared to other physiological domains due to limited public availability. Moreover, intrapartum recordings consist of only a small portion of the pretraining data, which may constrain labour-specific representation learning despite strong generalisation. Additionally, the model incorporates patient metadata, which may not be uniformly available across institutions and could limit applicability.

\section{Conclusion}
PRISM-CTG demonstrates that large-scale self-supervised pretraining can be effectively applied to CTG to learn transferable and generalisable in-domain representations. By incorporating patient metadata and engineered features as supervisory signals, the model learns clinically grounded representations that extend beyond purely morphological patterns. The results show consistent improvements over task-specific supervised models, stronger robustness to missing data, and improved cross-institutional generalisation. Collectively, our findings indicate that future work in automated CTG analysis should be driven by large-scale, clinically informed self-supervised pretraining as the primary training strategy, instead of continued reliance on narrow task-specific supervised optimisation.

\begin{credits}
\subsubsection{\ackname}
This research was supported by the UKRI Medical Research Council (MR/X029689/1)
\subsubsection{\discintname}
The authors have no competing interests to declare.

\end{credits}
%
%
%
\bibliographystyle{splncs04}
\bibliography{ECML_PKDD_2026_Author_Kit/mybibliography}

\end{document}